\documentclass[lettersize,journal]{IEEEtran}
\usepackage{amsmath,amsfonts}
\usepackage{algorithmic}
\usepackage{array}
\usepackage{textcomp}
\usepackage{stfloats}
\usepackage{url}
\usepackage{verbatim}
\def\BibTeX{{\rm B\kern-.05em{\sc i\kern-.025em b}\kern-.08em
    T\kern-.1667em\lower.7ex\hbox{E}\kern-.125emX}}
\usepackage{balance}

\usepackage{color}
\usepackage[utf8]{inputenc}
\usepackage{multirow}
\DeclareUnicodeCharacter{2212}{-}
\usepackage{booktabs}

\usepackage{times}
\usepackage{epsfig}
\usepackage{subcaption}
\usepackage{comment}
\usepackage{amsmath,graphicx}
\usepackage{xcolor}
\usepackage{amsmath}
\usepackage{amssymb}
\usepackage[accsupp]{axessibility}  %
\usepackage[pagebackref,breaklinks,colorlinks]{hyperref}
\usepackage{xcolor, colortbl}
\definecolor{almond}{RGB}{186,210,225}

\newcommand*{\our}{\texttt{SRMM}}

\newcommand\blfootnote[1]{%
  \begingroup
  \renewcommand\thefootnote{}\footnote{#1}%
  \addtocounter{footnote}{-1}%
  \endgroup
}

\begin{document}
\title{Modality Invariant Multimodal Learning to Handle Missing Modalities:
A Single-Branch Approach}
\author{Muhammad Saad Saeed$^{1\ast}$,  
Shah Nawaz$^{2\ast}$, 
Muhammad Zaigham Zaheer$^{3}$,
Muhammad Haris Khan$^{3}$, \\
Karthik Nandakumar$^{3}$,
Muhammad Haroon Yousaf$^{1}$, 
Hassan Sajjad$^{4}$,
Tom De Schepper$^{5}$,
Markus Schedl$^{2,7}$\\
$^{1}$Swarm Robotics Lab (SRL)-NCRA, University of Engineering and Technology Taxila,
$^{2}$Institute of Computational Perception, Johannes Kepler University Linz,
$^{3}$Mohamed Bin Zayed University of Artificial Intelligence,
$^{4}$Dalhousie University,
$^{5}$Interuniversity Microelectronics Centre (IMEC),\\
$^{6}$Human-centered AI Group, AI Lab, Linz Institute of Technology\\
\tt\small \{shah.nawaz, markus.schedl\}@jku.at, \{saad.saeed, haroon.yousaf\}@uettaxila.edu.pk, 
\tt\small \{muhammad.haris, zaigham.zaheer, karthik.nandakumar\}@mbzuai.ac.ae \\
\tt\small tom.deschepper@imec.be, \tt\small hsajjad@dal.ca
}

\maketitle

\begin{abstract}
Multimodal networks have demonstrated remarkable performance improvements over their unimodal counterparts.
Existing multimodal networks are designed in a multi-branch fashion that, due to the reliance on fusion strategies, exhibit deteriorated performance if one or more modalities are missing. 
In this work, we propose a modality invariant multimodal learning method, which is less susceptible to the impact of missing modalities. 
It consists of a single-branch network sharing weights across multiple modalities to learn inter-modality representations to maximize performance as well as robustness to missing modalities.
Extensive experiments are performed on four challenging datasets including textual-visual (UPMC Food-101, Hateful Memes, Ferramenta) and audio-visual modalities (VoxCeleb1). 
Our proposed method achieves superior performance when all modalities are present as well as in the case of missing modalities during training or testing compared to the existing state-of-the-art methods.
\end{abstract}


\section{Introduction}
\blfootnote{\textsuperscript{$\ast$}Equal contribution.}
\label{sec:intro}

\begin{figure*}
     \centering
     \begin{subfigure}[b]{0.21\textwidth}
         \centering
         \raisebox{0.25\height}{\includegraphics[width=\textwidth]{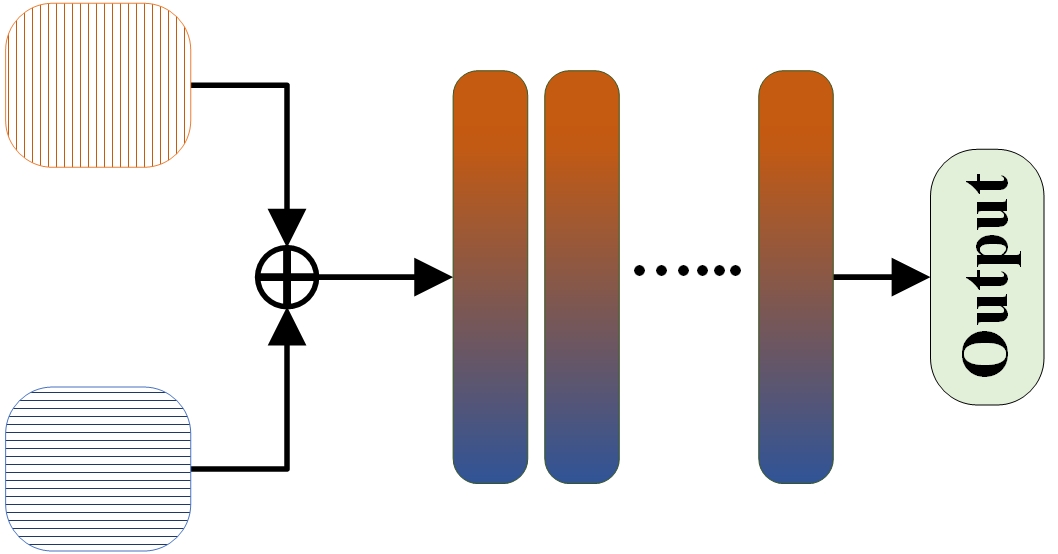}}
         \caption{Early fusion}
         \label{subfig:early}
     \end{subfigure}
     \hspace{0.03\textwidth}
     \begin{subfigure}[b]{0.21\textwidth}
         \centering
         \includegraphics[width=\textwidth]{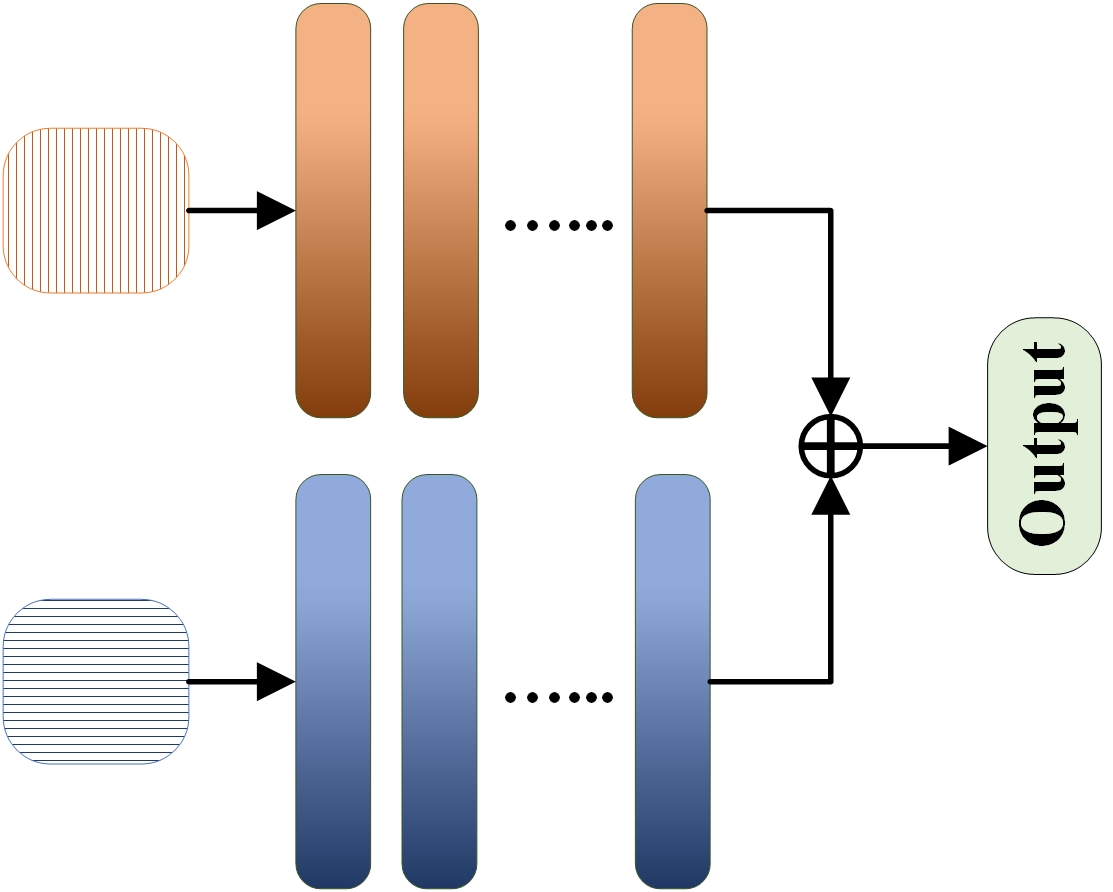}
         \caption{Late fusion}
         \label{subfig:late}
     \end{subfigure}
     \hspace{0.03\textwidth}
     \begin{subfigure}[b]{0.21\textwidth}
         \centering
         \includegraphics[width=\textwidth]{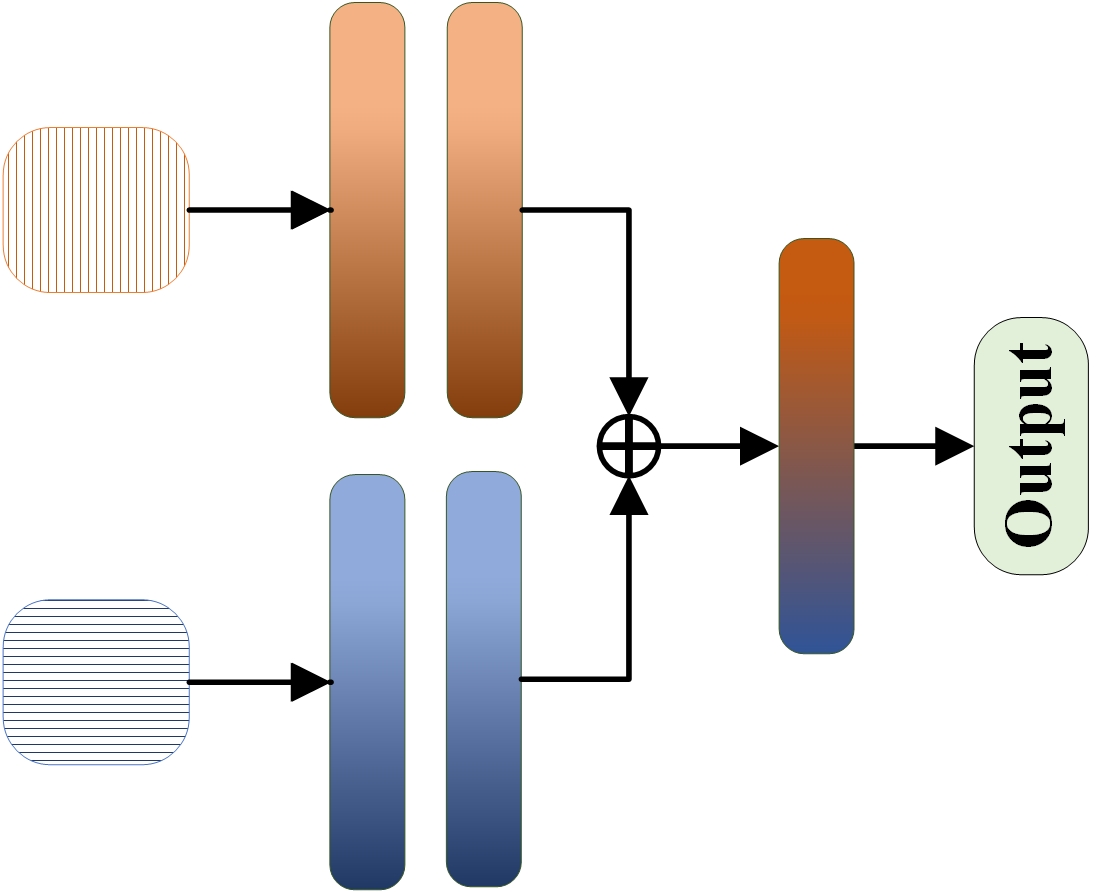}
         \caption{Middle fusion}
         \label{subfig:mid}
     \end{subfigure}
     \hspace{0.03\textwidth}     
     \begin{subfigure}[b]{0.21\textwidth}
         \centering
         \raisebox{0.25\height}
         {\includegraphics[width=\textwidth]{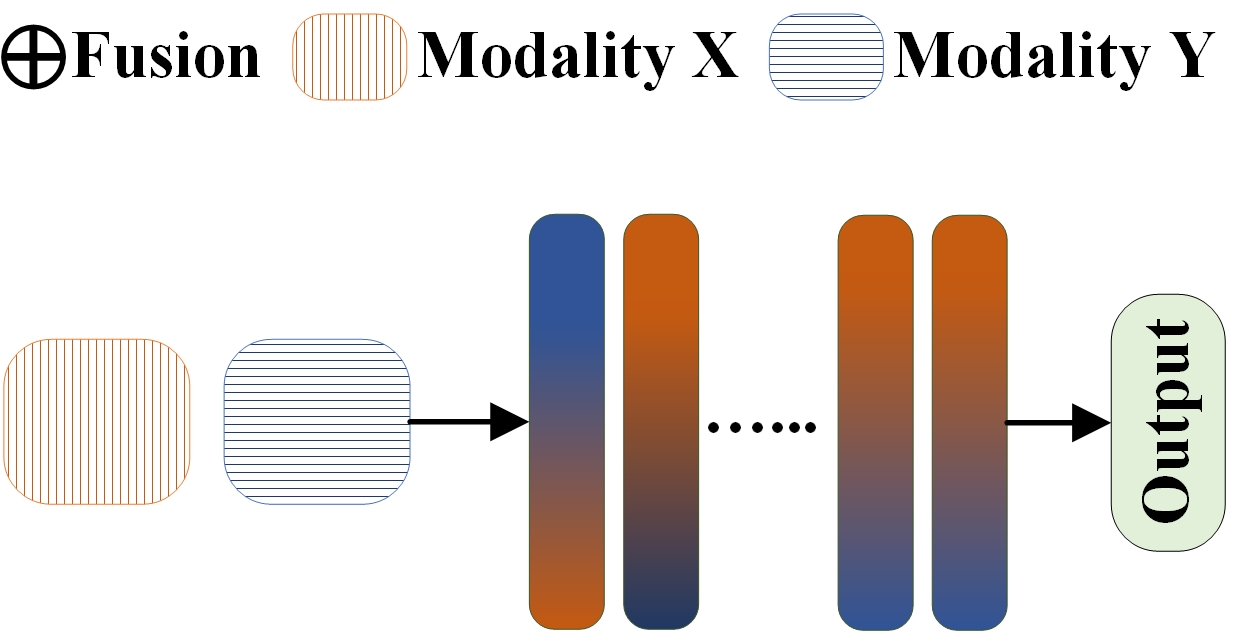}}
         \caption{Ours}
         \label{subfig:ours}
     \end{subfigure}
        \caption{
        Illustrations of commonly used multi-branch networks. These approaches learn a joint representation with fusion mechanisms (early, late or middle) from the embeddings of modality X and Y~\cite{feng2020deep}. In contrast, our proposed modality invariant method leverages only one branch to learn similar representations.}
        \label{fig:motivation}
\end{figure*}

\begin{table*}
\caption{Comparison of \our{} with  ViLT~\cite{kim2021vilt} on UPMC Food-$101$~\cite{wang2015recipe} dataset under different training and testing settings. $\Delta \downarrow$ indicates performance deterioration due to missing modality at test time. 
$\Delta \downarrow$  $=(P_{complete}-P_{missing})/P_{complete}$, where $P_{complete}$ and $P_{missing}$ are the performances over complete and missing modalities. 
Best results in each setting are shown in bold.
}
\centering
\footnotesize
\begin{tabular}{c|c|c|cc|cc|c|c}
\hline
\multirow{2}{*}{Dataset} & \multirow{2}{*}{Methods} & \multirow{2}{*}{Settings} & \multicolumn{2}{c|}{Training}  & \multicolumn{2}{c|}{Testing} & \multirow{2}{*}{Accuracy} & \multirow{2}{*}{$\Delta \downarrow$} \\
\cline{4-7}
& & &Image & Text & Image & Text &  & \\
\hline\hline
\multirow{4}{*}{UPMC Food-101} & \multirow{2}{*}{ViLT}  
                    & Complete Modalities & 100\%  & 100\% & 100\% & 100\% & 91.9 & -\\ 
                    &  &     Missing Modality   & 100\% & 100\% & 100\% & 30\% & 65.9 & 28.3\% \\

\cline{2-9}
                     & \multirow{2}{*}{\our}  
                     & Complete Modalities & 100\%  & 100\% & 100\% & 100\%   & \textbf{94.6} & -\\ 
                     & & Missing Modality  & 100\% & 100\% & 100\% & 30\% & \textbf{84.8 }& \textbf{12.3}\% \\
\hline
\end{tabular}

\label{tab:intro}
\end{table*}
\IEEEPARstart{M}{ultiple} modalities including text, image, video, and audio often contain complementary information about a common subject~\cite{baltruvsaitis2018multimodal,xu2023multimodal,zhang2024mmginpainting}. 
Different combinations of these modalities have been extensively studied to solve various tasks such as multimodal classification~\cite{kiela2018efficient}, cross-modal retrieval~\cite{wang2016learning,pang2015deep}, cross-modal verification~\cite{nagrani2018seeing}, multimodal named entity recognition~\cite{moon1078}, visual question answering~\cite{anderson2018bottom}, image captioning~\cite{vinyals2015show}, multimodal sentiment analysis~\cite{zhu2022multimodal}  and multimodal machine translation~\cite{elliott2016multi30k}. 
Multimodal modeling is challenging due to the difference in structure and representations of various modalities~\cite{baltruvsaitis2018multimodal}.  
The existing multimodal methods have commonly used neural network-based mappings to learn the joint representation of multiple modalities.
For example, separate independent networks are leveraged to extract embeddings of each modality to learn joint representations in multi-branch networks~\cite{reed2016learning, wang2016learning,faghri2018vse++,nagrani2018seeing,nagrani2018learnable,saeed2022fusion,kim2018learning}. 
Likewise, some recent multimodal methods have leveraged Transformers to learn joint representations using multi-branch networks~\cite{lu2019vilbert,tan2019lxmert}.
In these methods, the modular nature of the multi-branch networks is pivotal in developing various multimodal applications and have demonstrated remarkable performance over unimodal methods. 
However, a limitation of these methods is that they require complete modalities in training data to demonstrate good testing performance.

Multimodal data collected from the real-world are often imperfect due to missing modalities, resulting in a significantly deteriorated performance of the existing models~\cite{suo2019metric,ma2022multimodal,zhao2021missing,zeng2022tag, ma2021smil,lee2023multimodal,wang2023multi}. 
For example, as seen in Table \ref{tab:intro}, ViLT \cite{kim2021vilt}, a multimodal Transformer-based model, demonstrates a drop in performance of $28.3$\% when $70$\% of the text modality is missing at test time. 
This deteriorated performance renders multimodal learning ineffective for real-world scenarios where missing modality may be encountered.
The drop in performance may be attributed to the commonly used multi-branch design implementing fusion layer(s) for modality interaction (Figure \ref{fig:motivation}(a)-(c)). Such a design may learn weights in a way that the performance is highly dependent on the correct combination of input modalities~\cite{lu2019vilbert,kim2021vilt}.

In this work, we target the problem of robustness to missing modalities by hypothesizing that learning a shared representation across different modalities enables a common continuous representation space~\cite{firat-etal-2016-multi,multilingualMT_survey,liang2022foundations,ganhorsbnet}. 
Such inter-modality representations could possibly benefit in the case of missing modality. 
Motivated by this, we propose \textbf{S}ingle-branch approach \textbf{R}obust to \textbf{M}issing \textbf{M}odalities (\our{}) that utilizes weight sharing across multiple modalities in a single-branch network to enable learning of inter-modality representations (Fig. \ref{subfig:ours}). \our{} utilizes pre-trained embeddings of each modality and learns a joint representation using a modality switching mechanism to carry out the training.
It outperforms state-of-the-art (SOTA) methods on several multimodal datasets as well as demonstrates superior robustness against missing modality during training and testing.
For instance, as seen in Table~\ref{tab:intro}, compared to the existing multimodal SOTA method, ViLT~\cite{kim2021vilt}, \our{} results in a classification accuracy of $91.9$\% when both image and text modalities are completely available on the UPMC Food-$101$~\cite{wang2015recipe}. 
Under the same setting, our approach outperforms ViLT by achieving an accuracy of $94.6$\%.  
In case of severely missing modality (i.e., only $30$\% of text modality available during testing), ViLT demonstrates an accuracy of $65.9$\%.  In contrast, \our{} demonstrates substantial robustness against missing modality by achieving $84.9$\% accuracy when only $30$\% of textual modality is available, which is better than the unimodal performance.
Similar trends are observed across other multimodal classification datasets used to evaluate our approach (Section~\ref{subsec:missing}).
The key contributions of our work are as follows:
\begin{enumerate}
\itemsep0em 
  \item \our: A multimodal learning method robust to missing modalities during training and testing. 
  \item A modality invariant mechanism that enables weight sharing across multiple modalities in single-branch network.
  \item  A wide range of experiments are performed on challenging datasets including textual-visual (UPMC Food-101~\cite{wang2015recipe},
  Hateful Memes~\cite{kiela2020hateful}, and Ferramenta~\cite{gallo2017multimodal}) and audio-visual (Voxceleb$1$~\cite{nagrani2017voxceleb}) modalities. 
  \our{} exhibits SOTA performance when complete modalities are present. Similarly, in the case of missing modalities, our approach demonstrates superior robustness compared to existing SOTA methods.
\end{enumerate}


\textit{Difference from conference version:} 
A preliminary version of our single-branch training was published in the International Conference on Acoustics, Speech, and Signal Processing, 2023~\cite{saeed2023single}. In the preliminary version, we presented a single-branch network and compared its performance with existing multi-branch networks on a benchmark multimodal learning task using cross-modal verification and matching. 
The current work is a substantial extension of the conference paper with particular focus to demonstrate that single-branch network is robust to missing modality scenarios. 
First, we explore the robustness of single-branch networks on a new task of multimodal classification by utilizing four challenging multimodal datasets. Second, we extend the study to another modality pair (textual-visual) to study the generic applicability of single-branch training. Third, we extensively showcase that the single-branch training is significantly robust to missing modality.

\section{Related Work}
\label{sec:related-work}
The goal of multimodal learning is to leverage complementary information across multiple modalities to improve the performance of various machine learning tasks such as classification, retrieval, or verification. 
Each multimodal task is different from the other, while the underlying objective remains the same: to learn joint representations across multiple modalities~\cite{baltruvsaitis2018multimodal,xu2023multimodal,mai2019locally}. 
Existing multimodal methods employ multi-branch networks to learn joint representations by minimizing the distance between different modalities~\cite{wang2016learning,nagrani2018learnable,saeed2022fusion,kim2018learning,arevalo2017gated,vielzeuf2018centralnet,muennighoff2020vilio}.
Such methods using multi-branch networks have achieved remarkable performance~\cite{he2017fine,wang2015recipe,gallo2020image,arevalo2017gated,gallo2017multimodal,nawaz2019these}. 
However, most multimodal methods suffer from performance deterioration if some modalities become absent at test time, an issue often referred to as  missing modality problem~\cite{zhang2022m3care,wang2022m2r2,zhang2022multimodal,lin2023missmodal}. 

Considering the importance of multimodal methods, recent years have witnessed an increasing interest in handling the missing modality problem~\cite{ma2022multimodal,zhao2021missing,ma2021smil,lee2023multimodal,lin2023missmodal,ramazanova2024exploring,zeng2022robust,liaqat2024chameleon}.
Generally, existing multimodal methods that address this problem can be grouped into three categories. 
The first category is the input masking approach which randomly removes the inputs at training time to mimic missing modality information. 
For example, Parthasarathy et al.~\cite{parthasarathy2020training} introduced a strategy to randomly remove visual inputs during training  to mimic missing modality scenarios for the multimodal emotion recognition task.
The second category exploits the available modality to generate the missing one~\cite{ma2021smil,cai2018deep}. For example, Zhang et al.~\cite{zhang2022multimodal} generated the missing textual modality conditioned on the available visual modality. 
The third category learns a joint representation having related information from multiple modalities~\cite{wang2020transmodality}. 
For example, Han et al.~\cite{han2019implicit} learned audio-visual joint representations to improve the performance of the unimodal emotion recognition task, however, it cannot exploit complete modality information at the test time. 

In contrast to prior methods, we propose to learn modality invariant representations with a single-branch network employing weight sharing across multiple modalities.
\our{} not only demonstrates superior multimodal performance but also exhibits significant robustness towards missing modalities compared to the existing SOTA methods.

\begin{figure*}[t]
\centering
\includegraphics[scale=0.60]{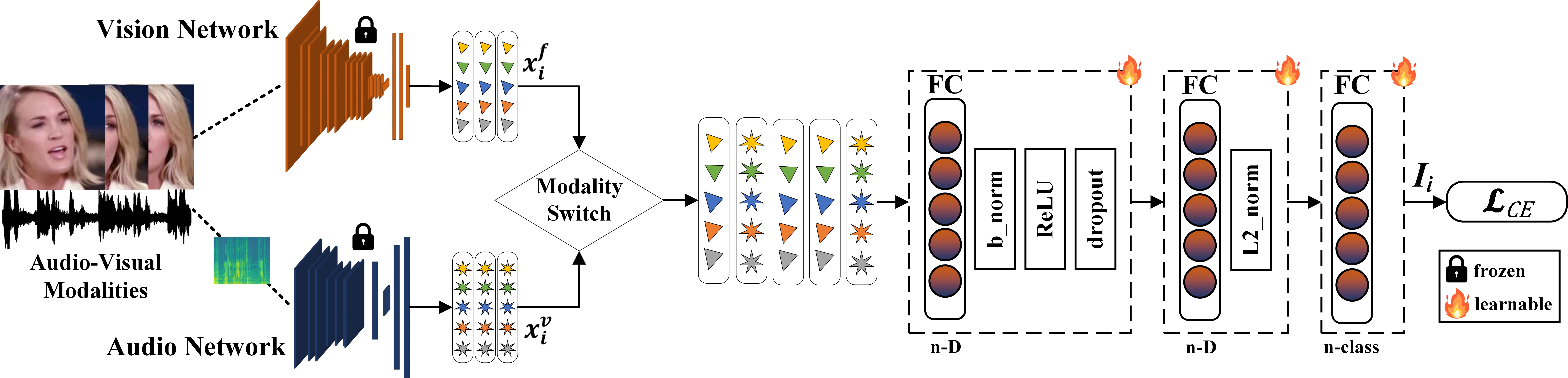}
  \caption{
  Overall architecture of \our{}. Modality-specific pre-trained networks (vision and audio networks in the given example) are used to extract embeddings which are passed through a modality switching mechanism and input to our single-branch network which learns modality independent representations to encode inter-modality representation with weight sharing across multiple modalities.
  }
\label{fig:main_arch}
\end{figure*}

\section{Methodology}
\label{section:overall_framework}
In this section, we describe \our{}, a multimodal learning method that is robust to missing modalities.
\our{} is built on the intuition that multiple embeddings extracted using modality-specific networks represent a similar concept but in a different representation space. The weight sharing using a single-branch network enables learning of inter-modality representations of these concepts. The model then benefits from the representations when a modality is missing at inference time. 
Fig.~\ref{fig:main_arch} presents our approach. 
In the following section, we explain modality embedding extraction,  single-branch network, and loss formulation used to train our network. 

\subsection{Preliminaries}
Given $\mathcal{D}=\{(M_{i}^{1},M_{i}^{2})\}_{i=1}^{N}$ is the training set where $N$ is the number of instances modality pairs $M_i^{1}$ and $M_i^{2}$.
Moreover, $x_{i}^{M_{i}^1}$ and $x_{i}^{M_{i}^2}$ are individual modality embeddings of the $i^{th}$ instance, respectively. 
Specifically, in the audio-visual case, the individual modality embeddings are represented as $x_{i}^{f}$ and $x_{i}^{v}$.
Moreover, each pair $(x_{i}^{f},x_{i}^{v})$ has a class label $y_{i}$. 
Multimodal learning aims at mapping multiple modalities into a common but discriminative joint embedding space where they are adequately aligned, and where instances from the same class are nearby while those from different classes are far apart~\cite{baltruvsaitis2018multimodal}.

\noindent\textbf{Limitations:} Typical existing multimodal methods take multiple modalities as input by using a fusion-centric multi-branch network ($\mathcal{C}_{MB}$) to perform the classification task:

\begin{equation}
    \tilde{y}_i = \mathcal{C}_{MB} (\{x_{i}^f, x_{i}^v\}, y_i)
    \label{eq:existingclassifiers}
\end{equation}

Such multi-branch configurations require modality-complete data to perform a given task and a missing modality results in significant performance deterioration~\cite{ma2022multimodal,suo2019metric,ma2021smil,liaqat2024chameleon}.

\subsection{\our{}}

\noindent\textbf{Motivation:} We hypothesize that learning a shared representation across different modalities enables a common continuous representation space. 
Such an inter-modality representation benefits in case of missing modality. 
Therefore, in our approach,  Eq.~\ref{eq:existingclassifiers} takes the following form:

\begin{equation}
    \tilde{y}_i = \mathcal{C}_{SB} (\{x_{i}^m\}, y_i)
    \label{eq:ours}
\end{equation}

\noindent where $x_{i}^m$ can be either $x_{i}^f$ or $x_{i}^v$.  
The mapping of multiple modalities into a common but discriminative joint embedding space enables the model to learn representations that are robust to a missing modality, resulting in a fusion-independent multimodal network.
We introduce a modality switching mechanism to determine the order in which embeddings of different modalities are input to the single-branch.
In our experiments (Section \ref{subsec:abl_switching_strategies}), unsurprisingly, we observed that our approach performs best when the input data is Independent and Identically Distributed (IID). However, surprisingly, other selections of modality switches yield comparable performance, demonstrating that our approach is generally robust to the order of the input modalities.

\noindent\textbf{Single-branch Network:} The overall architecture of \our{} is shown in Fig.~\ref{fig:main_arch}. The network comprises a single-branch of three blocks. The first block consists of a Fully Connected (FC) layer followed by Batch Normalization (b\_norm), ReLU, and dropout layers. The second block 
consists of an FC layer followed by normalization (L2\_norm) layer. The third block consists of an FC layer having the same size as the number of classes in a particular dataset followed by softmax. 
The weights of these FC layers are shared by different modalities embeddings which are input in a sequential fashion obtained from our modality-switching mechanism. 
At test time, if complete modalities are present, late fusion is employed by taking the average of the logits obtained from the softmax layer over all modalities. In the case of only one modality, the fusion mechanism is not employed. 

We employ cross entropy loss for training. 
Formally, we utilize a linear classifier with weights denoted as $\mathbf{W}=[\mathbf{w}_{1},\mathbf{w}_{2}, \dots,\mathbf{w}_{C}] \in \mathbb{R}^{d \times C}$ to compute the logits corresponding to $\mathbf{l}_{i}$ where $C$ is the number of classes and $d$ is the dimensionality of embeddings. The loss is then computed as:

\begin{equation}
    \mathcal{L}_{CE} = -log \frac{exp(\mathbf{l}_{i}^{T}\mathbf{w}_{y_{i}})}{\sum_{j=1}^{C}exp(\mathbf{l}_{i}^{T}\mathbf{w}_{j})}
\end{equation}

\our{} learns inter-modality representations across modalities, bringing them into a similar latent space. This results in a richer representation that is complemented by both modalities and yields a robust model when one of the modalities is missing. In other words, when a modality is missing, the available modality enables the use of the multimodal knowledge from the shared inter-modality representation.

\section {Experiments: Complete, Missing, and Corrupted Modalities}
\label{section:experi}
We evaluate \our{} on the multimodal classification task using four datasets including textual-visual modalities including UPMC Food-$101$~\cite{wang2015recipe}, Hateful Memes~\cite{kiela2021hateful}, Ferramenta~\cite{gallo2017multimodal} and audio-visual modalities based Voxceleb$1$~\cite{nagrani2017voxceleb}.
We conduct experiments using various settings including complete modalities and different levels of missing modalities during training and testing. 
Moreover, an extensive ablation study is performed to evaluate different design choices of our approach.
For a fair comparison, we adopt the same evaluation metrics used by the original authors of each dataset and the subsequent SOTA methods, i.e., classification accuracy and area under the receiver operating characteristic (AUROC).

  \begin{figure*}
      \centering
      \includegraphics[width=0.9\linewidth]{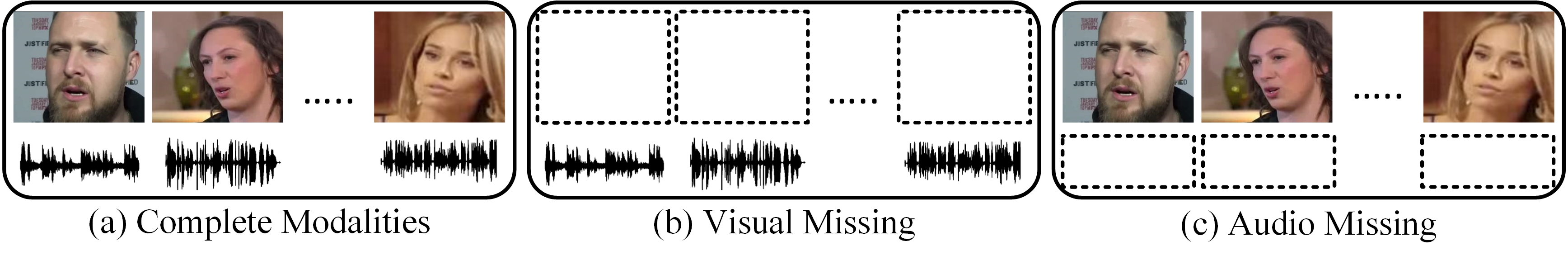}
      \caption{Examples of (a) complete modalities, (b) visual missing, and (c) audio missing settings.}
      \label{fig:example_dataset}
  \end{figure*}

\subsection{Datasets}
Recently, Ma et al.~\cite{ma2022multimodal} and Lee et al.\cite{lee2023multimodal} introduce an evaluation protocol to study the missing-modality problem during training and testing. 
To provide a comparison, we select the UPMC Food-$101$ and Hateful Memes
datasets from \cite{ma2022multimodal} and \cite{lee2023multimodal}.
In addition, we select an audio-visual dataset (VoxCeleb$1$) to evaluate the generic applicability of \our{} on other modalities.
Finally, we select a widely popular and challenging multimodal dataset, Ferramenta, which is curated to resolve ambiguities among visual samples by using the textual modality. 

\noindent \textbf{UPMC Food-$101$.}  It is a classification dataset consisting of textual and visual modalities. The dataset was crawled from the web and each entry consists of an image and the HTML web page on which it was found.
The dataset contains $90,704$ image-text pairs and $101$ classes, and is released with a pre-defined 75/25 train/test splits. \\
\noindent \textbf{Hateful Memes.} It is a multimodal dataset containing meme images and their respective textual contents as the two modalities, along with binary labels that identify the presence of hate speech in memes. 
The dataset contains $10,000$ memes. \\
\noindent \textbf{Ferramenta.} It consists of $88,010$ textual-visual pairs belonging to $52$ classes. The data is divided into $66,141$ instances for train and $2,186$ instances for test. \\
\noindent \textbf{Voxceleb$1$.} It is an audio-visual dataset of human speech videos of $1,251$ speakers, extracted `in the wild' from YouTube. The data is divided into $145,265$ instances for train and $8,251$ instances for test.

\noindent Fig.~\ref{fig:example_dataset} shows example of complete and missing visual and audio modalities.

\subsection{Implementation Details}

\noindent \textbf{Network Settings.} \our{} is trained using Adam optimizer with a learning rate of $0.01$ and dropout of $50\%$. The network has FC layers as: \{input\_dim, layer\_dim, layer\_dim, number of classes\}, where the input\_dim is $512$ for audio-visual and $768$ for textual-visual modalities. Moreover, layer\_dim is $2048$ for audio-visual and $768$ for textual-visual modalities.

\noindent \textbf{Modality-specific Embeddings.}
We employ modality-specific networks to extract embeddings as explained in this section. 
Additional analysis by using other modality-specific extractors is also provided in Section~\ref{subsec:emb_extractors} (Table \ref{tab:feats}). However, the following networks are our default experiment choices.

\noindent \textbf{Image Embeddings.}
We extract image embeddings using Contrastive Language–Image Pre-training (CLIP)~\cite{radford2021learning}. 
The size of the output embeddings is $768$, which matches with the corresponding text modality.\\
\noindent \textbf{Text Embeddings.}
We extract text embeddings from CLIP. The size of the output embedding is fixed to $768$ to match the corresponding image modality.\\
\noindent \textbf{Face Embeddings.}
We extract face embeddings using  Inception-ResNet-V$1$~\cite{szegedy2017inception} pre-trained with triplet loss~\cite{schroff2015facenet}. The size of output embeddings is $512$ which matches with the corresponding audio modality. \\
\noindent \textbf{Audio Embeddings.}
We extract audio embeddings using an utterance level aggregator~\cite{xie2019utterance} trained for a speaker recognition task with VoxCeleb$1$~\cite{nagrani2017voxceleb} dataset. The size of output embeddings is kept ($512$) to match the corresponding face embeddings.
The network is trained with a fixed-size spectrogram corresponding to a $2.5$ second temporal segment, extracted randomly from each utterance~\cite{xie2019utterance}.

\subsection{Evaluations Under Complete Modalities Setting} 
We first evaluate \our{} when complete modalities are present during training and testing and compare the results with existing SOTA methods. Tables~\ref{tab:sota-upmc},~\ref{tab:sota-hateful},~\ref{tab:sota-ferramenta}, and \ref{tab:voxceleb} present the results. 
\our{} achieves SOTA performance on three out of four datasets. 
More specifically, on the UPMC-Food-$101$ dataset (Table~\ref{tab:sota-upmc}), our model achieved a classification performance of $94.6$\%, outperforming all existing methods.
On Ferramenta and VoxCeleb$1$ (Table~\ref{tab:sota-ferramenta} and \ref{tab:voxceleb}) datasets, our model achieved $96.5$\% and $98.0$\% accuracy, respectively, outperforming all SOTA methods.
Only on the Hateful Memes dataset (Table~\ref{tab:sota-hateful}), \our{} did not achieve SOTA results but demonstrated a comparable performance with several methods except Vilio~\cite{muennighoff2020vilio}. It may be noted that Vilio is an ensemble method employing five vision and language models to achieve the reported AUROC, thus not directly comparable to any of the methods that are based on using a single system for inference.

\begin{table}[t]
\caption{Comparison of \our{} with state-of-the-art methods on UPMC-Food-$101$. Best results are shown in bold; second best are underlined.}
\centering
\footnotesize
\begin{tabular}{l|c}
\hline
Method   & Accuracy \\


\hline\hline
 
Wang et al.~\cite{wang2015recipe}   & 85.1 \\
\hline
Fused Representations~\cite{nawaz2018learning}   & 85.7 \\
\hline
 MMBT~\cite{kiela2019supervised}   & 92.1 \\
\hline
 BERT+LSTM~\cite{gallo2020image}  & 92.5 \\
\hline
 ViLT~\cite{kim2021vilt}     & 91.9 \\
\hline
Ma et al.~\cite{ma2022multimodal}   & 92.0 \\
\hline
TBN~\cite{saeed2022fusion}   & \underline{94.2} \\
\hline
\our{} & \textbf{94.6} \\

\hline
\end{tabular}

\label{tab:sota-upmc}
\end{table}

\begin{table}
\caption{Comparison of \our{} with state-of-the-art multimodal methods on Hateful Memes. 
$^\ast$ denotes the results from the Hateful Memes Challenge~\cite{kiela2021hateful}. 
\textsuperscript{\textdagger}An ensemble of five different vision and language models.
Best results are shown in bold; second best are underlined.}
\centering
\footnotesize
\begin{tabular}{l|c}
\hline
Method  & AUROC \\

\hline\hline
 
MMBT-Grid~\cite{kiela2019supervised}$\ast$      & 67.3 \\
\hline
MMBT-Region~\cite{kiela2019supervised}$\ast$    & 72.2 \\
\hline
ViLBERT~\cite{lu2019vilbert}$\ast$              & \underline{73.4} \\
\hline
Visual BERT~\cite{li2019simple}$\ast$           & 73.2 \\
\hline
ViLT~\cite{kim2021vilt}                    & 70.2 \\
\hline
Vilio~\cite{muennighoff2020vilio}\textsuperscript{\textdagger}        & \textbf{82.5} \\
\hline
\our{}                                            & 72.5 \\
\hline
\end{tabular}

\label{tab:sota-hateful}
\end{table}
\begin{table}
\caption{Comparison of \our{} with state-of-the-art multimodal methods on Ferramenta dataset.}
\centering
 \footnotesize
\begin{tabular}{c|c}
\hline
Method & Accuracy \\

 \hline\hline
Ferramenta~\cite{gallo2017multimodal}  & 92.9 \\
\hline
Fused Representations~\cite{nawaz2018learning}  &  94.8 \\
\hline
IeTF~\cite{gallo2018image}  & 95.2 \\
\hline
TBN~\cite{saeed2022fusion}    & 96.2 \\
\hline
MHFNet~\cite{yue2023multi}    & \underline{96.5} \\
\hline
\our{}   & \textbf{96.5} \\

\hline
\end{tabular}

\label{tab:sota-ferramenta}
\end{table}
\begin{table}
\caption{Comparison of \our{} with TBN on VoxCeleb$1$ dataset. 
}
\centering
 \footnotesize
\begin{tabular}{l|c}
\hline
Method  & Accuracy \\
 \hline\hline
TBN~\cite{saeed2022fusion}   & \underline{97.7} \\
\hline
\our{}  & \textbf{98.0} \\
\hline
\end{tabular}

\label{tab:voxceleb}
\end{table}

\begin{table*}
\caption{Evaluation of \our{} with different levels of available modality in test set using UPMC-Food-$101$ and Hateful memes datasets. 
Comparison is provided with ViLT~\cite{kim2021vilt}$^\ast$, Ma et al.~\cite{ma2022multimodal}, Kim et al.~\cite{kim2024missing}, and  TBN~\cite{saeed2022fusion}.
$^\ast$ViLT values are taken from \cite{ma2022multimodal}.
AUROC and accuracy are reported for Hateful Memes and UPMC-Food-$101$, respectively. 
Boldface and underline denote, respectively, the best and second best results.
}
\centering
\footnotesize
\begin{tabular}{c|cc|cc|c|c|c|c|c|c|c}
\hline
\multirow{2}{*}{Dataset} & \multicolumn{2}{c|}{Training}  & \multicolumn{2}{c|}{Testing} & \multirow{2}{*}{ViLT~\cite{kim2021vilt}} & \multirow{2}{*}{Ma et al.~\cite{ma2022multimodal}}& \multirow{2}{*}{Kim et al.~\cite{kim2024missing}} & \multicolumn{3}{c|}{TBN~\cite{saeed2022fusion}}  & \multirow{2}{*}{\our{}}  \\

\cline{2-5}
\cline{9-11}
&  Image & Text & Image & Text &  &  & &  Early & Mid & Late  & \\
 \hline\hline
\multirow{7}{*}{\rotatebox[origin=c]{90}{UPMC Food-101}}
                      & 100\% & 100\% & 100\% & 100\%     & 91.9   & 92.0 & 92.0 & \underline{94.5}  & 94.2  & 94.2 & \textbf{94.6} \\
                      & 100\% & 100\% & 100\% & 90\%      & 88.2   & 90.5 & - & 92.7  & 92.1  & \underline{93.0} & \textbf{93.2}  \\
                      & 100\% & 100\% & 100\% & 70\%      & 80.7   & 87.1 & 85.8 & \underline{89.4}  & 87.9  & 88.6 & \textbf{90.9}  \\
                      & 100\% & 100\% & 100\% & 50\%      & 73.3   & 82.6 & 82.1 & \underline{85.9}  & 83.6  & 84.3 & \textbf{88.2}    \\
                      & 100\% & 100\% & 100\% & 30\%      & 65.9   & 77.5 & 78.4 & \underline{81.8}  & 79.8  &79.9 & \textbf{84.8}     \\
                      & 100\%  & 100\% & 100\% & 10\%     & 58.4   & 73.3 & 74.7 & \underline{78.7}  & 75.5  & 75.4 & \textbf{83.3}     \\ 
                      & 100\%  & 100\% & 100\% & 0\%      & -   & -       & 72.6 & \underline{76.8}  & 73.3  & 73.1 & \textbf{82.0} \\ 
                      
\hline
\multirow{7}{*}{\rotatebox[origin=c]{90}{Hateful Memes}}                       
                      & 100\% & 100\% & 100\% & 100\%      & 70.2  & \underline{71.8} & 72.3 & 59.8 & 61.0  & 61.1 & \textbf{72.5} \\
                      & 100\% & 100\% & 100\% & 90\%      & 68.8   & \underline{69.7} & - & 59.6 & 60.7  & 60.9 &  \textbf{72.2}  \\
                      & 100\% & 100\% & 100\% & 70\%      & 65.9   & \underline{66.6} & 69.5 & 59.2 & 60.5  & 60.2 &  \textbf{72.0}   \\
                      & 100\% & 100\% & 100\% & 50\%      & 63.6   & \underline{63.9} & 67.4 & 59.1 & 60.4  & 60.0 &  \textbf{72.1}    \\
                      & 100\% & 100\% & 100\% & 30\%      & 60.2   & \underline{61.2} & 64.4 & 58.9 & 60.1  & 59.4 &  \textbf{71.3}     \\
                      & 100\%  & 100\% & 100\% & 10\%     & 58.0   & \underline{59.6} & 61.8 & 58.1 & 59.6  & 59.7 &  \textbf{71.2}     \\ 
                      & 100\%  & 100\% & 100\% & 0\%      & 54.9   & -                & 60.4 & 58.0 & 59.4  & \underline{59.5} & \textbf{71.2}   \\ 
\hline
\end{tabular}


\label{tab:percentage_missing}
\end{table*}

\begin{table}
\caption{Classification accuracy of \our{} on the configuration of 100\% missing modality in test set using Ferramenta and VoxCeleb$1$ datasets. Comparison of our approach is provided with the TBN \cite{saeed2022fusion} to understand the importance of our single branch design.
Best results are printed in boldface.}
\centering
\footnotesize
\resizebox{0.99\linewidth}{!}{
\begin{tabular}{c|cc|cc|c|c}
\hline
Dataset & \multicolumn{2}{c|}{Training}  & \multicolumn{2}{c|}{Testing}  & TBN~\cite{saeed2022fusion} & \our{} \\

\hline\hline
\multirow{4}{*}{\rotatebox[origin=c]{45}{Ferramenta}} 
                      & Image  & Text & Image & Text &                &    \\ 
                      \cline{2-5}
                      & 100\%  & 100\% & 100\% & 100\%   & \underline{96.2} & \textbf{96.5}  \\ 
                      & 100\% & 100\% & 100\%  & 0\%     & \underline{71.0}  & \textbf{92.3}  \\
                      & 100\% & 100\% & 0\%    & 100\%   & \underline{61.6} & \textbf{93.4} \\
\hline
\multirow{4}{*}{\rotatebox[origin=c]{45}{VoxCeleb$1$}}
                      & Image  & Audio & Image & Audio &                &    \\ 
                      \cline{2-5}
                      & 100\%  & 100\% & 100\%   & 100\%    & \underline{97.7}  & \textbf{98.0}  \\ 
                      & 100\%  & 100\%  & 100\%  & 0\%      & \underline{38.9}  & \textbf{84.7} \\
                      & 100\%  & 100\%  & 0\%    & 100\%    & \underline{31.5}  & \textbf{82.4} \\
\hline
\end{tabular}
}

\label{tab:all-data-single-branch}

\end{table}

\subsection{Evaluations Under Missing Modalities Setting}
\label{subsec:missing}
\noindent \textbf{Missing modalities during testing.} Ma et al.\cite{ma2022multimodal} have shown that multimodal methods are brittle to missing modalities at test time. \our{} aims to show better robustness towards missing modalities by learning intermodal representations.
Table~\ref{tab:percentage_missing} compares our approach with existing SOTA methods; ViLT~\cite{kim2021vilt}, Ma et al.~\cite{ma2022multimodal}, Kim et al.~\cite{kim2024missing} and Two-branch Network~\cite{saeed2022fusion} (TBN) for varying amounts of missing modality on UPMC Food-$101$ and Hateful Memes datasets.
In the manuscript, we consider a TBN as the baseline network to our proposed method as the input embeddings and loss formulations are the same while the network is comparable.
The only fundamental difference is that the TBN has a fusion component in a multi-branch design.
To provide extensive analysis, we consider several variants of the TBN including early, late, and middle fusion as seen in Fig.~\ref{fig:motivation}. We observed that the TBN with early fusion (similar to FOP~\cite{saeed2022fusion}) performs slightly better than its counterpart versions with late or middle fusion. Therefore, to avoid redundant results, we primarily use this network as a baseline in our manuscript for SOTA comparisons.
Overall, \our{} outperformed all existing SOTA methods including the TBN baselines with considerable margins. 
In the case of severely missing text modality (when only $10$\% is available), on the UPMC Food-$101$ dataset, \our{} demonstrates an accuracy of $83.3$\%.
Compared to this, TBN  with early fusion, ViLT, Ma et al, and Kim et al. demonstrate performances of $81.6\%$, $58.4\%$, $73.3\%$, and $74.7$\% respectively. Similarly, on the Hateful Memes dataset, \our{} demonstrates an AUROC of $71.2$\% when only $10$\% of text modality is available at test time. In comparison, TBN with early fusion, ViLT, Ma et al., and Kim et al.  demonstrate performances of $59.7\%$, $58.0\%$, and $59.6\%$, and $61.8\%$ respectively.
Similar trends are observed on Ferramenta and VoxCeleb$1$ datasets, as seen in Table \ref{tab:all-data-single-branch}, where comparisons are provided with the TBN with early fusion \cite{saeed2022fusion}. 
This demonstrates the significance of our proposed method for multimodal training robust to missing modalities. Our work serves as a proof of concept that encourages researchers to consider modality invariant learning in building robust multimodal methods.

\noindent \textbf{Missing modalities during training.} Recently, Lee et al.~\cite{lee2023multimodal} extended the evaluation protocol by introducing scenarios of missing modality during training and testing, i.e., $30$\% of a modality is available against $100$\% of the other modality at train and test time. 
Table~\ref{tab:reb_traing_missing} compares our approach with existing SOTA methods including ViLT~\cite{kim2021vilt}, Lee et al.~\cite{lee2023multimodal} on UPMC Food-$101$, and Hateful Memes datasets.
In most scenarios, \our{} demonstrates better performance than existing methods. 
As \our{} does not have a fusion-centric multi-branch design, it does not require modality-complete data during training and testing, demonstrating resilience to missing modality scenarios.

\begin{table*}
\caption{Comparison of \our{} with  ViLT~\cite{kim2021vilt}$^\ast$, Ma et al.~\cite{lee2023multimodal}, and Kim et al.~\cite{kim2024missing} on UPMC Food-$101$, and Hateful Memes under different training and testing settings. 
$^\ast$ViLT values are taken from \cite{lee2023multimodal}.
Best results in each setting are shown in bold.
}
\centering
\footnotesize
\begin{tabular}{c|cc|cc|c|c|c|c}
\hline
\multirow{2}{*}{Dataset}  & \multicolumn{2}{c|}{Training}  & \multicolumn{2}{c|}{Testing} & \multirow{2}{*}{ViLT~\cite{kim2021vilt}}  & \multirow{2}{*}{Lee et al.\cite{lee2023multimodal}} & \multirow{2}{*}{Kim et al.\cite{kim2024missing}} & \multirow{2}{*}{\our{}} \\
\cline{2-5}
 &Image & Text & Image & Text &   &   & &\\
\hline\hline
\multirow{3}{*}{UPMC Food-$101$} & 100\%  & 100\%  & 100\% & 100\%  & 91.9  & \underline{92.0} & \underline{
92.0}  & \textbf{94.6} \\ 
                                 & 100\%  & 30\%  & 100\% & 30\%   & 66.3  & 74.5  & \underline{78.8}  & \textbf{84.8}  \\
                                 & 30\%   & 100\%  & 30\% & 100\%  & 76.7  & \underline{86.2}     & \textbf{86.9}  & 79.9 \\
                                 
                                 \hline
\multirow{3}{*}{Hateful Memes}   & 100\%  & 100\%  & 100\% & 100\% & 70.2 & 71.0               & \underline{72.3}  & \textbf{72.5}\\   
                                 & 100\%  & 30\%  & 100\% & 30\% & 60.8  & 59.1                            & \underline{61.4} &  \textbf{72.1} \\ 
                                 & 30\%   & 100\%  & 30\% & 100\%  & 61.6  & \underline{63.1}     & \textbf{66.5}  & 58.8 \\ 
\hline
\end{tabular}

\label{tab:reb_traing_missing}
\end{table*}

\subsection{Evaluation Under Corrupted Modalities}
In real-world scenarios, it is often possible that a modality is not missing but corrupted due to several reasons including faulty equipment, low bandwidth, etc. In order to evaluate the robustness of \our{} on corrupted modalities, we perform a series of experiments by adding different levels of noise to the feature values of both modalities in the test set and report the results in Table \ref{tab:noise}. As seen, \our{} shows reasonable tolerance to 100\% corrupted modalities with several levels of Gaussian noise ($\mu=0, \sigma = {0.1, 0.5, 1.0}$) at test time. Under extreme noise ($\mu=0, \sigma = {1.0}$), the performance however deteriorates noticeably.

\begin{table}
\centering
\footnotesize
\resizebox{0.99\linewidth}{!}{
\begin{tabular}{c|c|cc|cc|c}
\hline
\multirow{2}{*}{Dataset} & \multirow{2}{*}{$\sigma$} & \multicolumn{2}{c}{Training}  & \multicolumn{2}{c|}{Testing} & \multirow{2}{*}{Acc.}  \\

\cline{3-6}
& & Image & Text & Image & Text &    \\
 \hline\hline
\multirow{4}{*}{\rotatebox[origin=c]{0}{Food-101}} & 0.0
                      & 100\%  & 100\% & 100\% & 100\% & \textbf{94.6}  \\ 
 & 0.1
                      & 100\%  & 100\% & 100\% & 100\% & 94.4   \\ 
& 0.5
                      & 100\%  & 100\% & 100\% & 100\% & 86.8  \\

& 1.0
                      & 100\%  & 100\% & 100\% & 100\% & 49.8   \\ 
\hline
\end{tabular}
}
\caption{The performance of our approach with $100$\% corrupted modalities by adding Gaussian noise. The standard deviation ($\sigma$) is varied whereas mean ($\mu$) is set to $0$. 
Results are reported on accuracy with best results shown in boldface.
}
\label{tab:noise}
\end{table}

\begin{table*}
\caption{Performance comparison of \our{} with the extracted embeddings using various pre-trained models. Best results are obtained when using CLIP as an image and text feature extractor.
}
\centering
 \footnotesize
\begin{tabular}{c|c|c|c|c|cc|cc|c}
\hline
\multirow{2}{*}{Dataset} & \multirow{2}{*}{Config.} &  \multirow{2}{*}{Image Emb.} & \multirow{2}{*}{Text Emb.} & \multirow{2}{*}{Emb Size} & \multicolumn{2}{c|}{Training}  & \multicolumn{2}{c|}{Testing} & \multirow{2}{*}{Accuracy} \\

\cline{6-9}
 &  &  & & & Image & Text & Image & Text &   \\
\hline\hline
\multirow{10}{*}{\rotatebox[origin=c]{90}{UPMC Food-101}} & \multirow{5}{*}{\rotatebox[origin=c]{90}{Unimodal}} & ResNet-$101$ & -  &  2048    & 100\% &  - & 100\% & - & 57.8 \\
                      & & ViT & - &  768     & 100\% &  - & 100\% & - & 69.5 \\
                      & & CLIP & - &  768    & 100\% &  - & 100\% & - & 81.7 \\
                      & & - & Doc$2$Vec &  300    & - &  100\% & - & 100\% & 74.1 \\
                      & & - & CLIP &  768    & - &  100\% & - & 100\% & 86.2\\
                      \cline{2-10}
                      & \multirow{5}{*}{\rotatebox[origin=c]{90}{Multimodal}} & ResNet-$101$ & Doc$2$Vec  &  2048    & 100\% &  100\% & 100\% & 100\% & 87.5 \\ 
                      & & ViT          & Doc$2$Vec  &  768     & 100\% & 100\% & 100\% & 100\% & 88.6  \\
                      & & CLIP         & Doc$2$Vec  &  768     & 100\% & 100\% & 100\% & 100\% & 92.2  \\
                      & & ViT          & CLIP       &  768     & 100\% & 100\% & 100\% & 100\% & 92.8  \\
                      & & CLIP         & CLIP       &  768     & 100\% & 100\% & 100\% & 100\% & \textbf{94.6}  \\
                      
\hline
\end{tabular}

\label{tab:feats}
\end{table*}

\section{Additional Experiments}
In this section, we provide further analysis on the impact of various embedding extractors and different modality switching strategies on the training and robustness of \our{}.

\begin{figure}[t]
    \centering
    \includegraphics[width=0.99\linewidth]{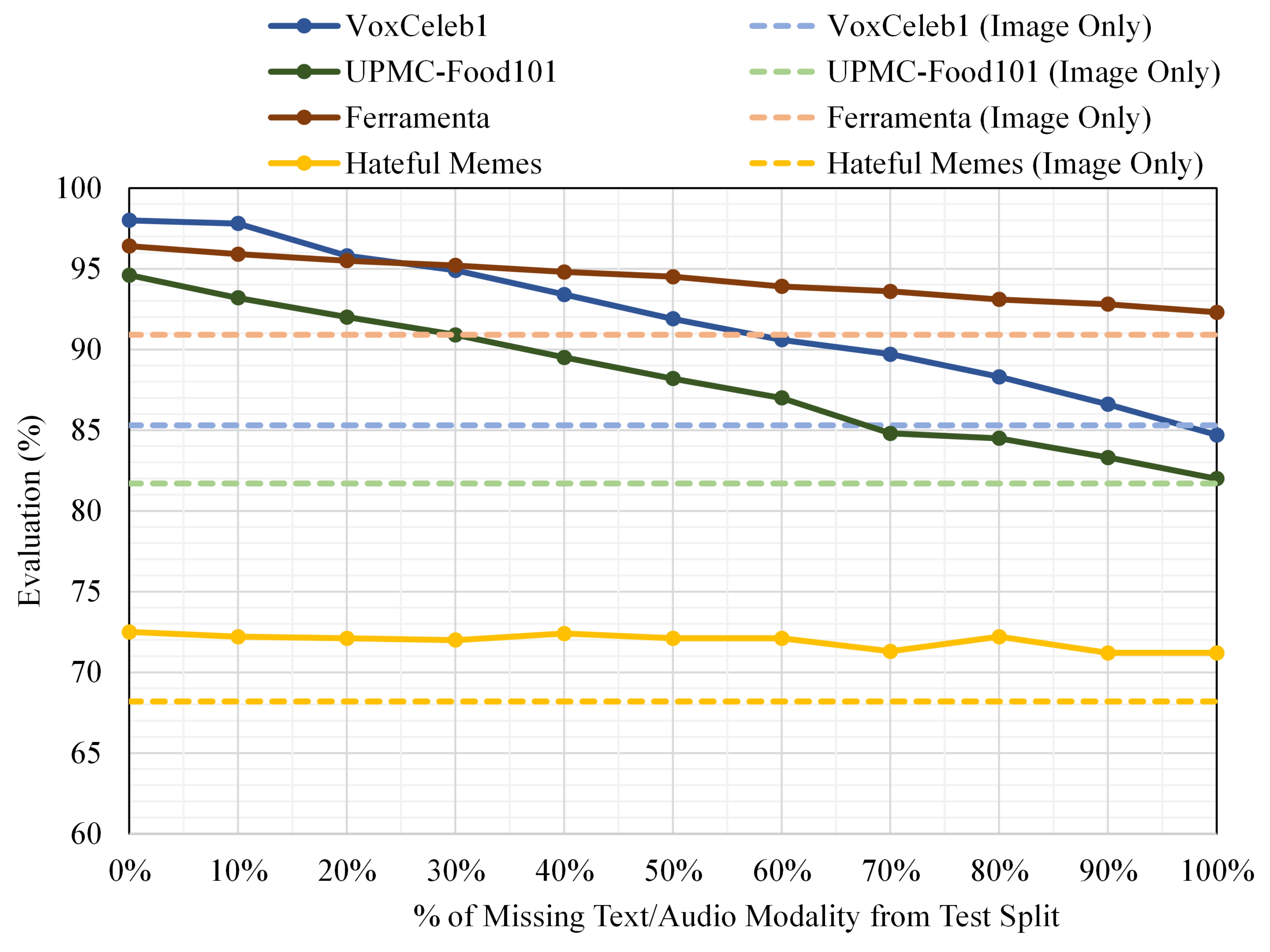}
    \caption{Performance evaluation of \our{} on missing modalities over four datasets including textual-visual (UPMC Food-101, Hateful Memes, Ferramenta) and audio-visual modalities (VoxCeleb$1$). Dotted lines represent unimodal results. Audio modality (in case of VoxCeleb$1$) and text modality (in case of other three datasets) is gradually dropped  from 0\% to 100\% by randomly eliminating samples from the test data.}
    \label{fig:percentage_missing_modalities}
\end{figure}

\begin{table}

\centering
\caption{Performance analysis of the  modality switching strategies.
Results are reported using 100$\%$ modalities in the training set.
Best results are shown in boldface.}
\resizebox{0.99\linewidth}{!}{
\footnotesize
\begin{tabular}{c|c|cc|c|c}
\hline
\multirow{2}{*}{Dataset} & \multirow{2}{*}{Strategy}  & \multicolumn{2}{c|}{Testing} & \multirow{2}{*}{Accuracy}   &\multirow{2}{*}{$\Delta \downarrow$}\\

\cline{3-4}
  &  & Image & Text & &\\
\hline\hline
\multirow{6}{*}{\rotatebox[origin=c]{0}{Food-101}} & \multirow{2}{*}{S-1}  & 100\% &  100\%  & \textbf{94.6} & -  \\
&   & 100\% &  0\%    & \textbf{82.0} & \textbf{13.3\%}  \\
\cline{2-6}
&\multirow{2}{*}{S-2}  & 100\% &  100\%  & 94.5  & -   \\
&                &       100\% &  0\%   &  81.1  &  14.3\%  \\
\cline{2-6}
&\multirow{2}{*}{S-3}  & 100\% &  100\%  & 93.2 & -  \\
&                     & 100\% &  0\%    & 81.7 & 13.6\%  \\
\hline
\end{tabular}

}

\label{tab:switching}
\end{table}

\subsection{Multimodal vs. Unimodal Training}
Generally, multimodal networks are popular because of their performance improvements over the models trained on individual modalities. However, when a multimodal method is exposed to missing modalities, the performance often drops lower than that of the models trained on individual modalities, voiding the benefit of multimodal training~\cite{ma2022multimodal}. Therefore, we performed a series of experiments to observe whether our approach demonstrates better performance in case of missing modality compared with training on individual modalities only. To carry out these experiments, instances from the test data are randomly eliminated to evaluate the robustness of the proposed method on missing modalities.
Fig.~\ref{fig:percentage_missing_modalities} shows that our approach demonstrates less susceptibility as the percentage of missing data increases for either of the modalities. Notably, compared to the models trained and tested on individual modalities, our approach retains better performance even when one of the modalities is missing over $90$\%. 
This demonstrates the significance of our proposed single-branch network in multimodal learning.

\subsection{Embedding Extractors}
\label{subsec:emb_extractors}
We carry out experiments using different pre-trained models as feature extractors including ResNet-$101$~\cite{he2016deep}, ViT~\cite{dosovitskiy2020image}, and CLIP~\cite{radford2021learning} for image embeddings and Doc2Vec~\cite{le2014distributed} and CLIP for text embeddings. 
The experiments were conducted in unimodal settings as well as in complete modality settings during training and testing. 

Overall, any combination of embeddings in multimodal training of \our{} outperforms the unimodal setting. For example, ResNet-101 and Doc2Vec multimodal training yields an accuracy of 87.5\% which is significantly higher than the unimodal performances of these embedding extractors, i.e., 57.8\% and 74.1\%. 
Similarly, using CLIP embeddings, \our{} improves upon the performance of unimodal training by a significant margin of 8.4\% in the case of text modality and 12.9\% in the case of image modality.

\subsection{Modality Switching Strategies}
\label{subsec:abl_switching_strategies}
We devise and study the impact of various switching strategies:
\begin{enumerate}
\itemsep0em 
\item[\texttt{S-1}] Randomly selecting either of the available modalities resulting in a multimodal embedding stream at the output of the switch. In this strategy, all batches are multimodal and batch selection is also random. 
\item[\texttt{S-2}] In each epoch, 50\% batches are multimodal as discussed in the first strategy, while the remaining 50\% batches are unimodal. For each unimodal batch, either of the modalities is randomly selected. During training, the batch selection is random, resulting in a mixed stream of unimodal and multimodal batches.
\item[\texttt{S-3}] In this strategy, all batches are unimodal. For each batch, either of the modality is randomly selected. During training, unimodal batches are then randomly selected, resulting in a multi-modal stream of unimodal batches.  
\end{enumerate}

\noindent Table~\ref{tab:switching} compares the results on the UPMC Food-$101$ dataset. \texttt{S-1} resulted in the best performance over the three strategies studied in this section by demonstrating 94.6\% accuracy and a drop of 13.3\% when the text modality is 100\% missing during testing. 
On the other hand, \texttt{S-2} resulted in a slightly lower accuracy of 94.5\% and a drop of 14.3\% when the text modality is completely missing. 
\texttt{S-3} demonstrates the lowest performance with only 93.2\% accuracy on complete modalities and drops by 13.6\% when the text modality is entirely missing. 
Interestingly, all selections of modality switches yield comparable performance, demonstrating that our approach is generally robust to the order of the input modalities. Overall, as \texttt{S-1} performs comparatively better, the results reported in our manuscript use it as default.

\subsection{Qualitative Results}
\label{subsec:tsne}
\begin{figure*}[t]
\centering
\includegraphics[width=0.85\linewidth]{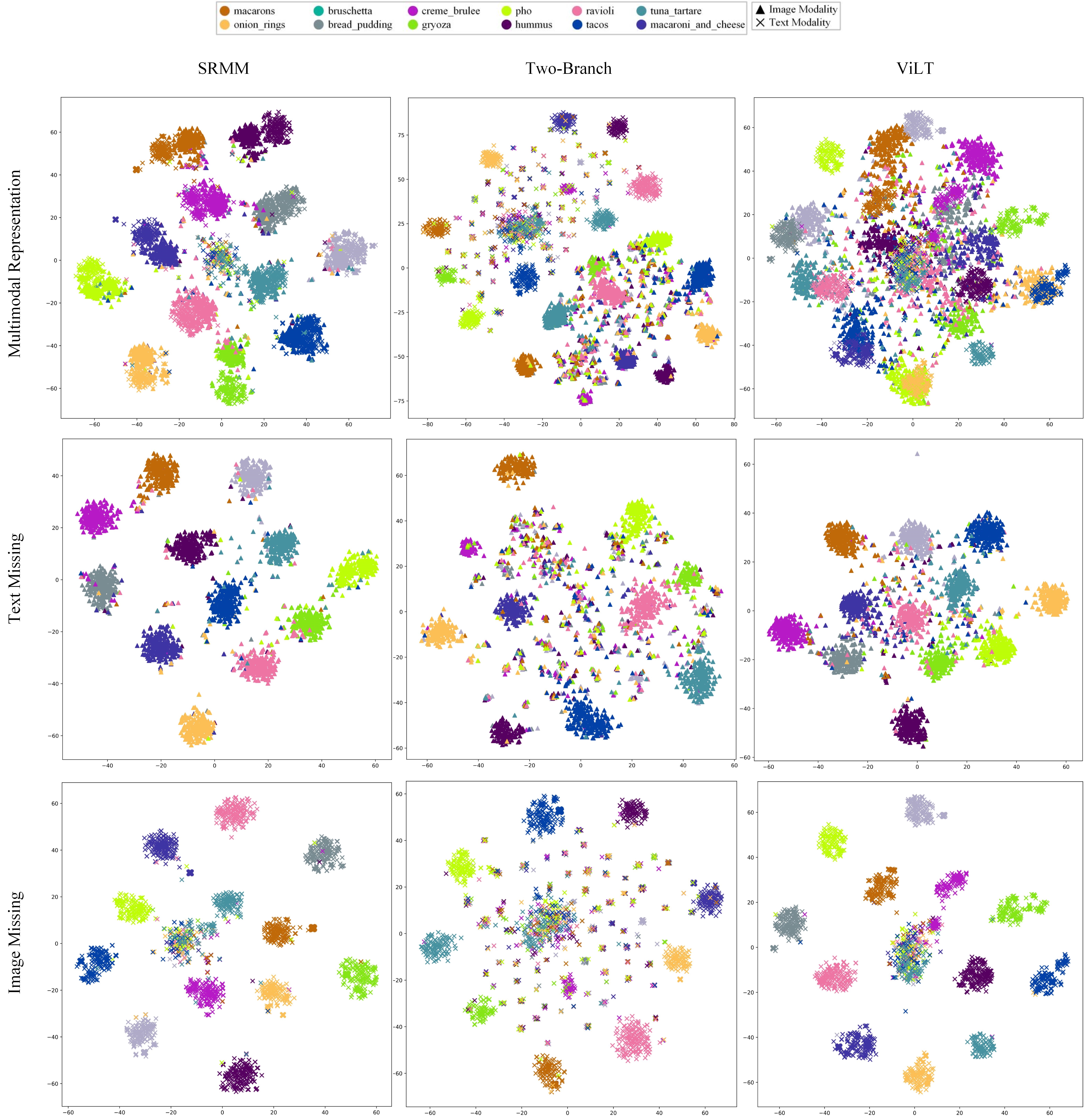}
\caption{
t-SNE visualizations of the embedding space of \our{} (embeddings from the second block), ViLT and TBN on test set of UPMC Food-$101$.
It can be seen that \our{} not only enhances the classification boundaries when complete modalities are available at test time but also retains these boundaries when the textual modality is completely missing during test time. 
}
\label{fig:ntsne}
\end{figure*}

In addition to the empirical results and analysis, we use t-SNE to visualize the embedding space of \our{} with complete modalities as well as missing textual and visual modality on UPMC Food-$101$. 
The visualizations are helpful in observing the overall effect of \our{} compared with the baseline (TBN) and ViLT.
Fig.~\ref{fig:ntsne} shows the multimodal embeddings extracted from the second block of \our{} along with ViLT and TBN. 
It can be seen that \our{} improves the overall classification boundaries highlighting the success of \our{}. 
In addition, Fig.~\ref{fig:ntsne} shows the embeddings extracted from \our{} when textual and visual modality is completely missing at test time. 
Although some distortions are noticeable, the overall separability of the classes is retained compared to ViLT and TBN
This demonstrates the robustness of \our{} towards missing modalities. 

\section{Advantages and Limitations}
\noindent\textbf{Advantages:~} 
Notable advantages of our approach are high performance and robustness to missing modality. Moreover, \our{} includes a significantly smaller network that results in fewer parameters for training. 
For example, compared to $2.44$ million parameters of TBN~\cite{saeed2022fusion}, \our{} requires merely $1.26$ million parameters when both are trained on image and text modalities. This makes training of \our{} more efficient compared to Transformers or other complex attention-based mechanisms \cite{ma2022multimodal,lee2023multimodal}. 

\noindent\textbf{Limitations:~} \our{} utilizes a modality switching mechanism leveraging weight sharing across multiple modalities with a single-branch network. Such a design requires extracted representations from modality-specific networks to have the same embedding size. 
We can perform a transformation of the embedding to the required size. However, this requires further experimentation and is out of the scope of our work.

\section{Conclusion}
\label{sec:conclusion}
We proposed a modality invariant multimodal learning method that is substantially robust against missing modalities during training and testing compared to existing methods. 
To ensure modality invariant learning, a modality switching mechanism serializes the embedding streams to a single-branch network. It facilitates complete weight sharing of the network and encodes the shared semantics across modalities. 
Extensive experiments are performed on four datasets including audio-visual (VoxCeleb$1$) and textual-visual modalities (UPMC Food-$101$, Hateful Memes, and Ferramenta).  The proposed method is thoroughly evaluated for missing modalities.  
The performance drop of the proposed network is noticeably smaller than the existing methods showing significant robustness.

\bibliographystyle{IEEEbib}
\bibliography{IEEEbib}


 
\vspace{11pt}

\begin{IEEEbiography}[{\includegraphics[width=1in,height=1.25in,clip,keepaspectratio]{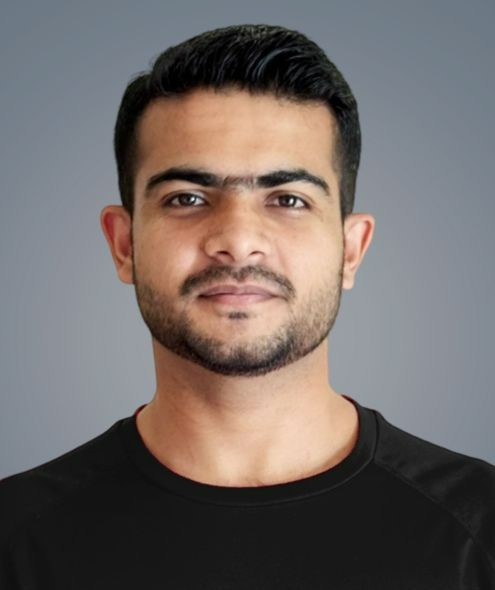}}]{Muhammad Saad Saeed} is working as a Research Associate at Swarm Robotics Lab under National Centre of Robotics and Automation Pakistan. His research interests are focused on multimodal learning, AI-on-the-Edge, Speech, and Audio Processing. He is the Member of IEEE and SPS.
\end{IEEEbiography}

\begin{IEEEbiography}[{\includegraphics[width=1in,height=1.25in,clip,keepaspectratio]{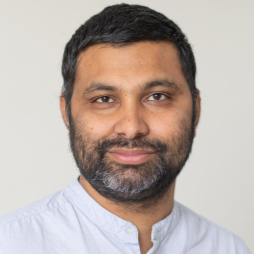}}]{Shah Nawaz} is an assistant professor at Johannes Kepler University Linz, Austria. He received bachelor degree in computer engineering from University of Engineering \& Technology, Taxila Pakistan and master degree in embedding systems from Technical University of Eindhoven, Netherlands, and PhD degree in computer science from University of Insubria, Italy. His research interests are focused on multimodal representation learning.
\end{IEEEbiography}

\begin{IEEEbiography}[{\includegraphics[width=1in,height=1.25in,clip,keepaspectratio]{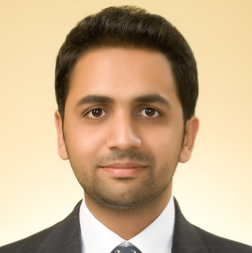}}]{Muhammad Zaigham Zaheer} is currently associated with Mohamed bin Zayed University of Artificial Intelligence as a Research Fellow. He received his Ph.D. degree from the University of Science and Technology, South Korea, in 2022. His current research interests include anomaly detection in images/videos, federated/collaborative learning, multimodal learning, data augmentation, and minimally supervised training.
\end{IEEEbiography}

\begin{IEEEbiography}[{\includegraphics[width=1in,height=1.25in,clip,keepaspectratio]{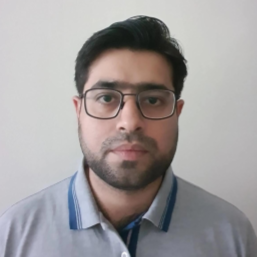}}]{Muhammad Haris Khan}  received the PhD degree in
computer vision from the University of Nottingham, U.K. He is a faculty member with the Mohamed bin Zayed University of Artificial Intelligence, UAE. Prior to MBZUAI, He was research scientist with the Inception Institute of Artificial Intelligence, UAE. He has published several papers in top computer vision venues. His research interests span active topics in computer vision.
\end{IEEEbiography}
\begin{IEEEbiography}[{\includegraphics[width=1in,height=1.25in,clip,keepaspectratio]{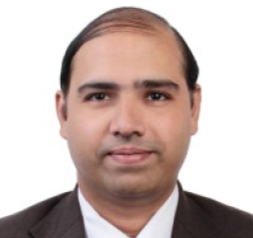}}]{Karthik Nandakumar} is an associate professor of computer vision at Mohamed bin Zayed University of Artificial Intelligence. His primary research interests include computer vision, machine learning, biometric recognition, applied cryptography, and blockchain. Prior to joining Mohamed bin Zayed University of Artificial Intelligence, Nandakumar was a research staff member at IBM Research – Singapore from 2014 to 2020 and a scientist at the Institute for Infocomm Research, A*STAR, Singapore from 2008 to 2014..
\end{IEEEbiography}
\begin{IEEEbiography}[{\includegraphics[width=1in,height=1.25in,clip,keepaspectratio]{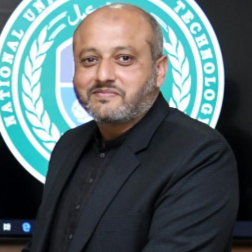}}]{Muhammad Haroon Yousaf} (Senior Member, IEEE) is working as a Professor/Chairman of Computer Engineering at University of Engineering and Technology Taxila, Pakistan. His research interests are Image Processing, Computer Vision, and Robotics. He is also the Director of Swarm Robotics Lab under National Centre of Robotics and Automation Pakistan. He has published more than 80 papers in International SCI-Indexed Journals/Conferences. He has secured many funded (Govt. and Industry) research projects in his areas of interest. Prof. Haroon has received the Best University Teacher Award by Higher Education Commission Pakistan in 2014. He is also the member of IEEE Signal Processing Society. He is also serving as Associate Editor of IEEE Transactions on Circuits and Systems on Video Technology (TCSVT). He is also appointed as Professor Extraordinarius at University of South Africa since February 2023.
\end{IEEEbiography}
\begin{IEEEbiography}[{\includegraphics[width=1in,height=1.25in,clip,keepaspectratio]{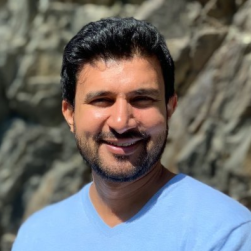}}]{Hassan Sajjad} is an Associate Professor in the Faculty of Computer Science at Dalhousie University, Canada, and the director of the HyperMatrix lab. His research focuses on natural language processing (NLP) and safe and trustworthy AI.
Dr. Sajjad has done pioneering work on interpretability which is recognized at several prestigious venues such as NeurIPS, ICLR, AAAI and ACL and is featured in prominent tech blogs including MIT News. 
\end{IEEEbiography}
\begin{IEEEbiography}[{\includegraphics[width=1in,height=1.25in,clip,keepaspectratio]{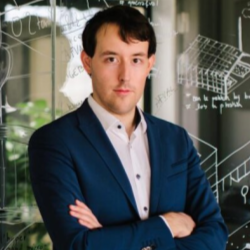}}]{Tom De Schepper} received its PhD in Computer Science in September 2019 at the University of Antwerp. He currently works as a Principle Member of Technical Staff in the AI \& Algorithms department of IMEC, Belgium. His current research situates on the intersection of AI and sensing. He is also a visiting professor at the department of Computer Science at University of Antwerp. He is co-author of more than 50 peer-reviewed articles.
\end{IEEEbiography}
\begin{IEEEbiography}[{\includegraphics[width=1in,height=1.25in,clip,keepaspectratio]{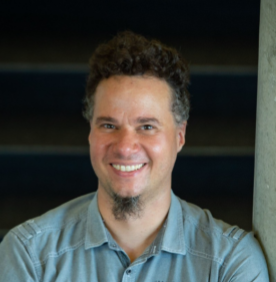}}]{Markus Schedl} is a full professor at the Johannes Kepler University Linz / Institute of Computational Perception, leading the Multimedia Mining and Search group. In addition, he is head of the Human-centered AI group at the Linz Institute of Technology (LIT) AI Lab. He graduated in Computer Science from the Vienna University of Technology and earned his Ph.D. from the Johannes Kepler University Linz. Markus further studied International Business Administration at the Vienna University of Economics and Business Administration as well as at the Handelshögskolan of the University of Gothenburg, which led to a Master's degree. His main research interests include recommender systems, information retrieval, natural language processing, multimedia, machine learning, and web mining. He (co-)authored more than 250 refereed articles in journals and conference proceedings as well as several book chapters. 
\end{IEEEbiography}
%


\vfill

\end{document}